\documentclass{article}

\pdfoutput=1

\usepackage[final,nonatbib]{nips_2018}

\usepackage[utf8]{inputenc} 
\usepackage[T1]{fontenc}    
\usepackage{hyperref}       
\usepackage{url}            
\usepackage{booktabs}       
\usepackage{amsfonts}       
\usepackage{nicefrac}       
\usepackage{microtype}      

\usepackage{fancyhdr,graphicx,amsmath,amssymb}

\title{High Quality Protein Q8 Secondary Structure Prediction by Diverse Neural Network Architectures}

\author{
  Iddo Drori\\
  Columbia University\\
  \texttt{idrori@cs.columbia.edu} \\
  \And
  Isht Dwivedi \\
  Columbia University \\
  \texttt{isht.dwivedi@columbia.edu} \\
  \And
  Pranav Shrestha \\
  Columbia University \\
  \texttt{ps2958@columbia.edu} \\
  \And
  Jeffrey Wan \\
  Columbia University \\
  \texttt{jw3468@columbia.edu} \\
  \And
  Yueqi Wang \\
  Columbia University \\
  \texttt{yw3169@columbia.edu} \\
  \And
  Yunchu He \\
  Columbia University \\
  \texttt{yh3050@columbia.edu} \\
  \And
  Anthony Mazza \\
  Columbia University \\
  \texttt{am4564@columbia.edu} \\
    \And
  Hugh Krogh-Freeman \\
  Columbia University \\
  \texttt{hk2903@columbia.edu} \\
    \And
  Dimitri Leggas \\
  Columbia University \\
  \texttt{ddl2133@columbia.edu} \\
    \And
  Kendal Sandridge \\
  Columbia University \\
  \texttt{ks3311@columbia.edu} \\
      \And
  Linyong Nan \\
  Columbia University \\
  \texttt{ln2401@columbia.edu} \\
    \And
  Kaveri Thakoor \\
  Columbia University \\
  \texttt{kat2193@columbia.edu} \\
   \And
  Chinmay Joshi \\
  Columbia University \\
  \texttt{caj2163@columbia.edu} \\
    \And
  Sonam Goenka \\
  Columbia University \\
  \texttt{sg3625@columbia.edu} \\
    \And
  Chen Keasar \\
  Ben Gurion University \\
  \texttt{keasar@cs.bgu.ac.il} \\
    \And
  Itsik Pe'er \\
  Columbia University \\
  \texttt{itsik@cs.columbia.edu} \\
}

\begin{document}

\maketitle

\begin{abstract}
We tackle the problem of protein secondary structure prediction using a common task framework. This lead to the introduction of multiple ideas for neural architectures based on state of the art building blocks, used in this task for the first time. We take a principled machine learning approach, which provides genuine, unbiased performance measures, correcting longstanding errors in the application domain. We focus on the Q8 resolution of secondary structure, an active area for continuously improving methods. We use an ensemble of strong predictors to achieve accuracy of 70.7\% (on the CB513 test set using the CB6133filtered training set). These results are statistically indistinguishable from those of the top existing predictors. In the spirit of reproducible research we make our data, models and code available \cite{drori18ssp}\footnote{Codebase: \url{https://github.com/idrori/cu-ssp}}, aiming to set a gold standard for purity of training and testing sets. Such good practices lower entry barriers to this domain and facilitate reproducible, extendable research.
\end{abstract}

\section{Introduction}
\label{introduction}

Proteins are the major building blocks of life on earth, and the mediators of almost all chemical and biophysical events in living organisms. They are polymer chains of amino acid residues, whose sequences (aka primary structure) dictate stable spatial conformations, known as the native structures. These structures in turn enable the biological functions of proteins. The sequence space of proteins is vast, 20 possible residues per position, and evolution has been sampling it over billions of years. Thus, current proteins are highly diverse in sequences, structures and functions. Predicting the 3D structure of a protein (PSP) from its linear sequence of amino acid units is a fundamental problem in computational biology, which is open for 50 years already. Virtually all the diverse approaches to PSP use, as their stepping stone, a prediction of the protein's secondary structure, the focus of the current study.

\begin{figure}[ht]
\begin{minipage}[b]{0.49\linewidth}
\centering
\includegraphics[width=0.6\linewidth]{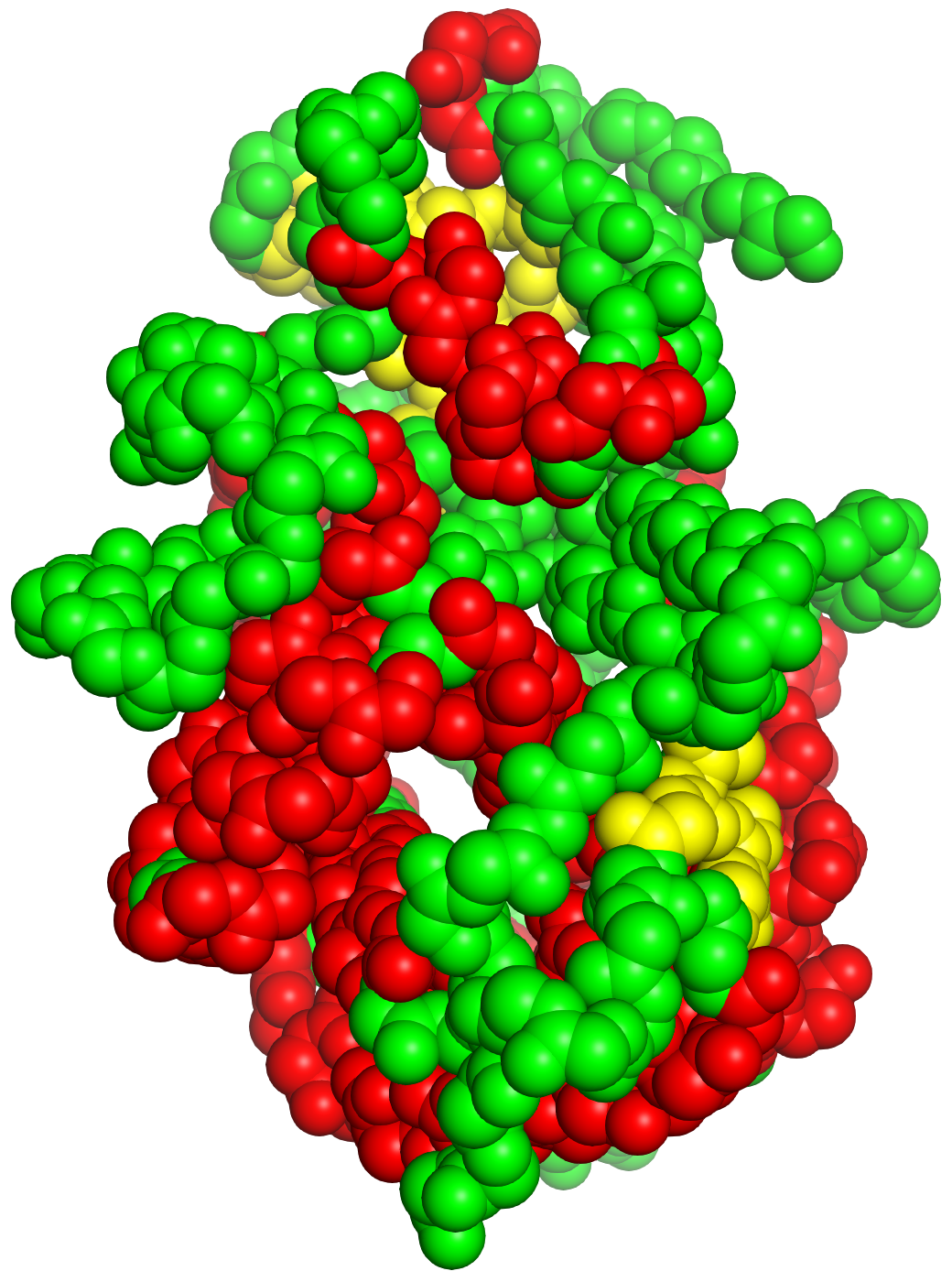}
\label{fig:q3}
\end{minipage}
\begin{minipage}[b]{0.49\linewidth}
\centering
\includegraphics[width=0.6\linewidth]{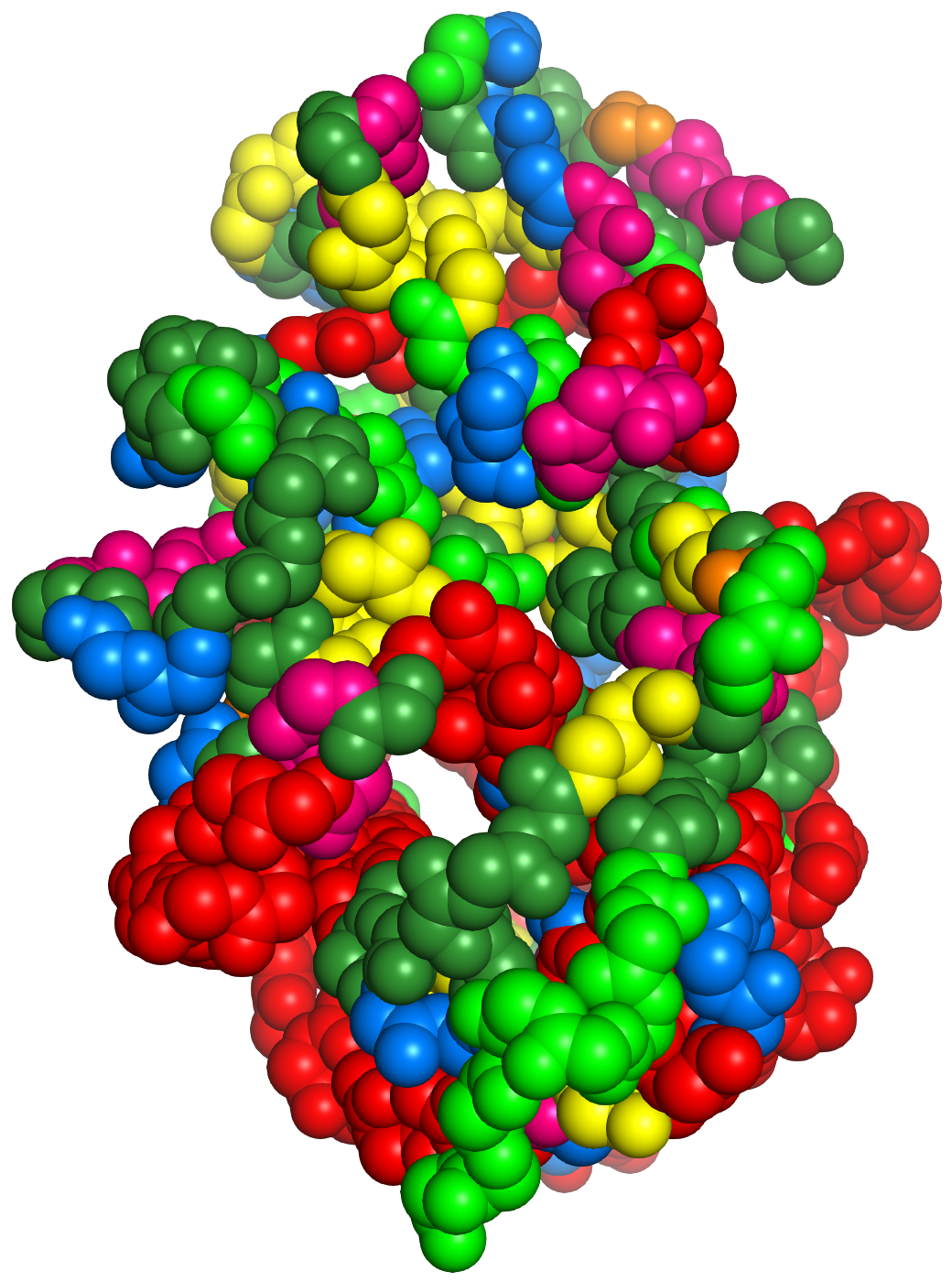}
\label{fig:q8}
\end{minipage}
\caption{Q3 (left); Q8 (right) secondary structure spheres for protein 1AKD in CB513 dataset.}
\label{fig:proteinsq3q8}
\end{figure}

Underneath the high diversity of protein structures, lies a relatively small set of recurrent patterns of torsion angles and hydrogen bonds that allow the protein to accommodate both local ({\it i.e.,} close chain positions) and non-local constraints. These patterns, which are known as secondary structure elements, imply the classification of the protein's residues to a relatively small number of structural classes known as the secondary structure. Since the mid-80s the dictionary of secondary structure patterns (DSSP) that suggested eight such classes has become the gold standard of the field \cite{kabsch1983dictionary}. Figure \ref{fig:proteinsq3q8} show a protein residues as spheres colored by their Q3 and Q8 structures. As secondary structure elements are stabilized by both local and non-local interactions, the tendency of protein segments to adopt them is sequence dependent. Beta-strand, for example, is a common pattern that implies a stretch of residues of the "E" (extended) class. It is characterized by alternating hydrophobic (oil-like) and hydrophilic (water-loving) residues. Such correspondence between two alphabets calls for the development of prediction methods, and indeed as early as the mid 70s secondary structure prediction (SSP) has gained much interest and was tackled by a wide variety of statistical approaches \cite{chou1974prediction, garnier1978analysis, deleage1987algorithm, holley1989protein, gibrat1987further, kneller1990improvements}. To ease the prediction challenge, these studies typically merged the eight DSSP classes to only three. In the early 90s Rost and Sander \cite{rost1993prediction, rost1994combining}, augmented protein sequences by profiles, derived from multiple sequence alignment of homologous proteins. They also introduced multi-tier neural networks and with these advances reached a landmark success of over 70\% accuracy in the three-state prediction scheme (Q3), dramatically outperforming previous approaches. Their success paved the way to further studies that provided more elaborate implementations of these concepts \cite{jones1999protein, zemla2003lga, wilson2004improved, midic2005improving, dor2007achieving, magnan2014sspro, heffernan2015improving, wang2016raptorx, busia2017next}, increasing the success rate of Q3 up to 84\%. However, as performance approached the postulated theoretical limit (85\%-88\%) \cite{Yang2018Sixty}, interest in the problem declined and progress became negligible over almost a decade. 
Recently however, interest has rekindled, as scholars replaced the relatively modest goal of predicting three classes by the more ambitious prediction of eight classes (Q8) \cite{wang2010Protein8class}. In the past five years there has been a steady and slow improvement in Q8 secondary structure prediction accuracy using deep neural networks \cite{zhou2014deep, sonderby2014protein, wang2016protein, li2016protein, busia2016protein, johansen2017deep, busia2017next, fang2018mufold}. This work reports the integration of multiple ideas for improving Q8 secondary structure prediction using an ensemble of predictors to achieve state of the art accuracy on the CB513 \cite{james1999evaluation} test set using a small training set of with less than 20\% identity of sub-sequences. 

In 2014 Zhou and Troyanskaya published their GSN method \cite{zhou2014deep} for SSP. To evaluate the performance of their method, they created a new benchmark termed CB6133 \cite{zhou14cb6133}. They used homologs-filtered subset of CB6133 to reach a then-record Q8 accuracy of 66.4\% for CB513 \cite{james1999evaluation} data-set. They also split CB6133 to training, validation and test sets and reported 72.1\% Q8 accuracy. They made their benchmark publicly available in accessible numpy format, stirring a wave of studies and publications, including the current one. In comparison, we have reached state of the art accuracy for CB513 of 70.5\% and best known published accuracy of 76.3\% for CB6133. We report data issues with the CB6133 standard that we ran into. These were since quickly fixed by the responsive authors on 10.28.18. We report results of a common task challenge, predicting Q8 secondary structures using novel network architectures. We report evaluation of these models and their ensemble on the CB513 data-set, reaching accuracy equivalent to current top predictors. In the spirit of reproducible research we make our data, models, and code fully available.

\section{Data and training}
\label{data}

\subsection{CB6133 dataset: correcting a long standing error}
We began this work by using the CB6133 dataset \cite{zhou2014deep} with the same train, validation, and test splits as used by other work for comparison \cite{zhou14cb6133}. We achieved the best known published performance on this dataset using the same published splits \cite{zhou14cb6133} as shown in Table \ref{tab:cb6133comparisonwithothers}. 
However, unfortunately, while using CB6133 dataset we couldn't but notice that it includes duplicate entries and the training, validation and test sets were not strictly disjoint. As a result of our finding the dataset splits were corrected by their creators and the valid splits re-published online by the 2014 authors on 10.28.18 \cite{zhou14cb6133}. Our contribution clears a long standing error in the field.

\subsection{Training data used for testing CB513: setting up standards}
We use the CB513 dataset \cite{james1999evaluation} for testing which is valid, does not contain any duplicates, and is disjoint from the training set we use CB6133filtered (after removing duplicates). Recent work \cite{zhang2018prediction} performs a comprehensive performance comparison on this test set, however uses different training sets of different sizes as if they were the same, and therefore we do not report those results here. To standardize our comparison and minimize redundancy we  considered the smaller training dataset, CB6133filtered, which multiple methods have in common. We achieve state of the art results as shown in Table \ref{tab:cb513comparisonwithothers}. 

\subsection{Features and output classes}   
Following Zhou and Troyanskaya \cite{zhou2014deep} we use 46 features per residue \textit{i.e., sequence positions} to classify each residue to one of nine classes. A subset of 22 features represent residue type by one-hot encoding. In addition to the standard 20 residue types: A, C, E, D, G, F, I, H, K, M, L, N, Q, P, S, R, T, W, V, and Y, we use X for non-standard residues (\textit{e.g., selenium methionine}) and noSeq for padding. A second subset of 22 features represent residue's position in a position specific substitution matrix (PSSM aka profile) that was generated by PSI-BLAST \cite{Altschul1997Gapped}. Again the last two features represent non-standard residues and padding. Finally, two binary features indicate the first and last position of the sequence. All sequences are padded with one-hot encoding of noSeq to length 700. The output classes include the eight classes defined by DSSP \cite{kabsch1983dictionary}: L, B, E, G, I, H, S, and T.

\section{Methods}
\label{methods}
The neural network architectures of our 6 models are diverse. This section provides a detailed description and an illustration of each architecture. The training time for each of the models is around one hour using an Nvidia 1080 GPU.

\subsection{Bidirectional LSTMs with attention}
\begin{figure*}[ht]
\centering
\includegraphics[width=1\textwidth]{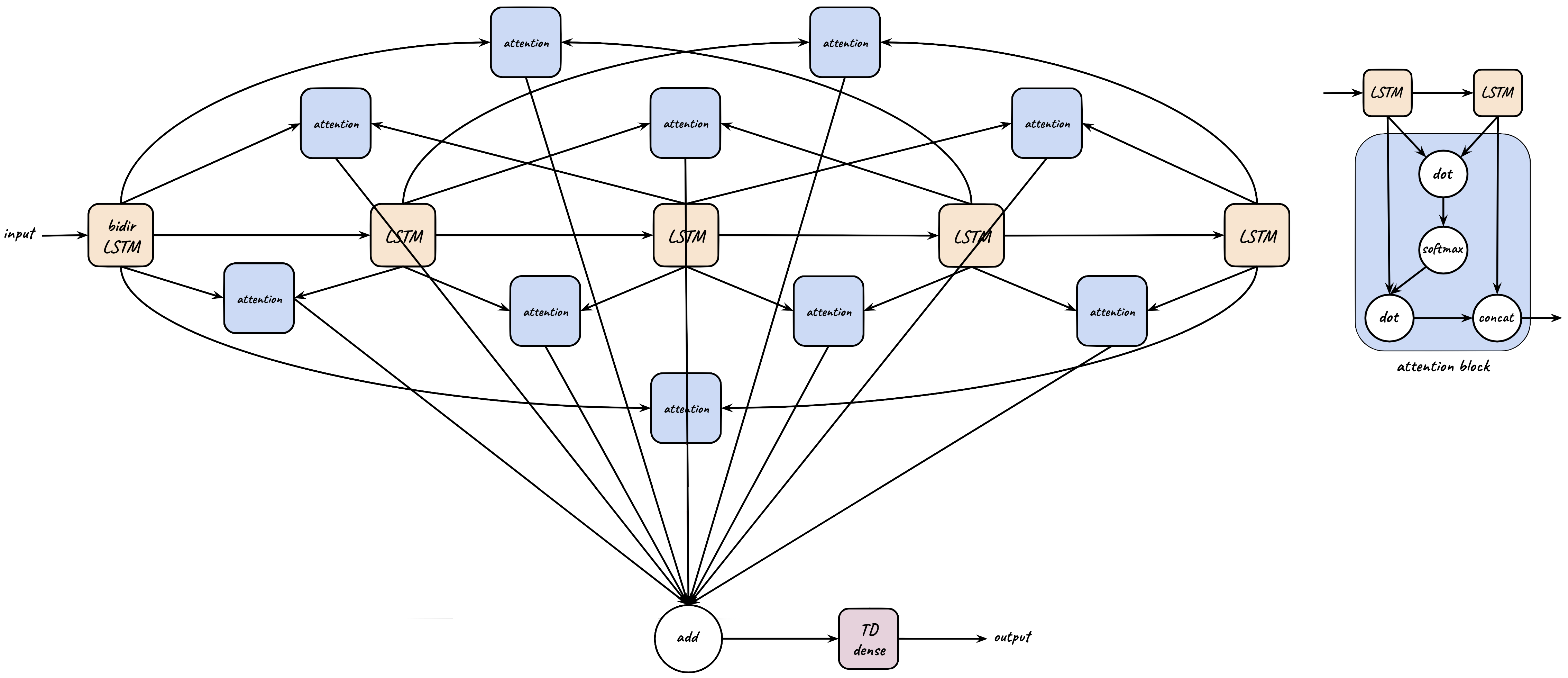}
\caption{Bidirectional LSTMs with attention.}
\label{fig:model5}
\end{figure*}

 Figure \ref{fig:model5} shows the architecture for this model. An embedding of the bigram amino acid sequence input is concatenated with the profile features and passed to a bidirectional LSTM (with 75 units), followed by 4 unidirectional LSTMs (each with 150 units). The initial state of the latter LSTM is initialized by the last hidden state of the former LSTM (concatenated in the case of biLSTM). For each possible pair of LSTMs, an attention mechanism \cite{luong2015effective} is applied using output of the latter LSTM as queries and output of the former LSTM as keys and values. This process generates 10 attention outputs, which are then added and passed to two fully-connected layers. This is the first time an attention mechanism \cite{luong2015effective,vaswani2017attention} is used for this problem achieving state of the art results without using convolutions.

\vspace{-5pt}

\vspace{-10pt}

\subsection{U-Net with convolution blocks}
\begin{figure*}[ht]
\centering
\includegraphics[width=1\textwidth]{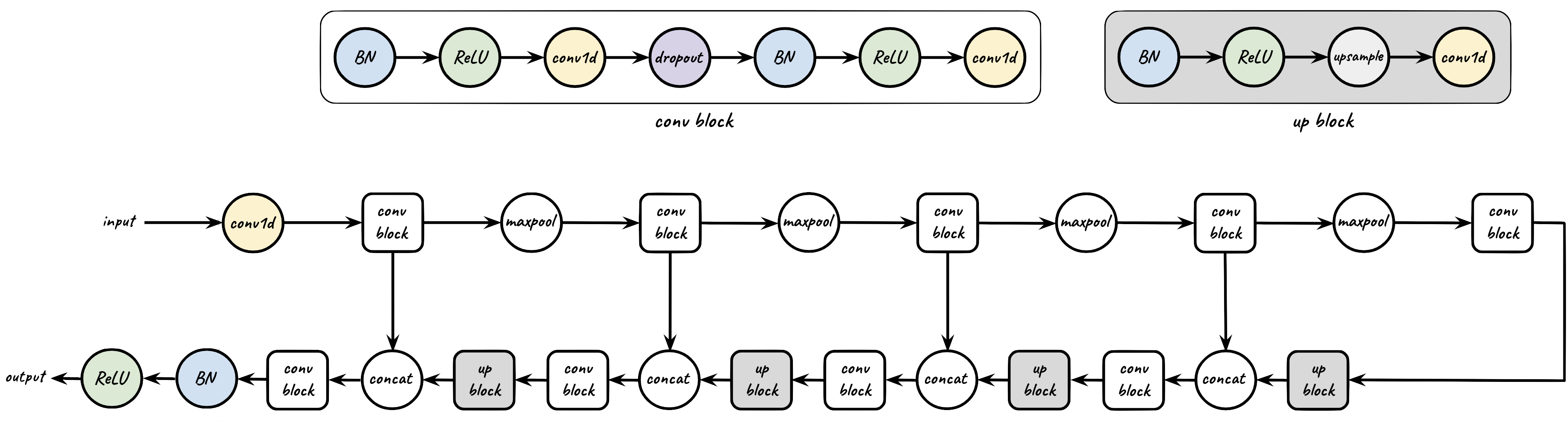}
\caption{U-Net with convolution blocks.}
\label{fig:model2}
\end{figure*}

Figure \ref{fig:model2} shows the architecture for this model. A fully convolutional model, using a one-dimensional U-Net \cite{ronneberger2015u} with dropout \cite{srivastava2014dropout} and batch normalization \cite{ioffe2015batch}. The profile input matrix is concatenated with the output of the embedding layer and fed into the first layer of a 1D U-Net. 

\vspace{-5pt}

\vspace{-10pt}

\subsection{Bidirectional GRU with convolutional blocks}
\begin{figure*}[ht]
\centering
\includegraphics[width=1\textwidth]{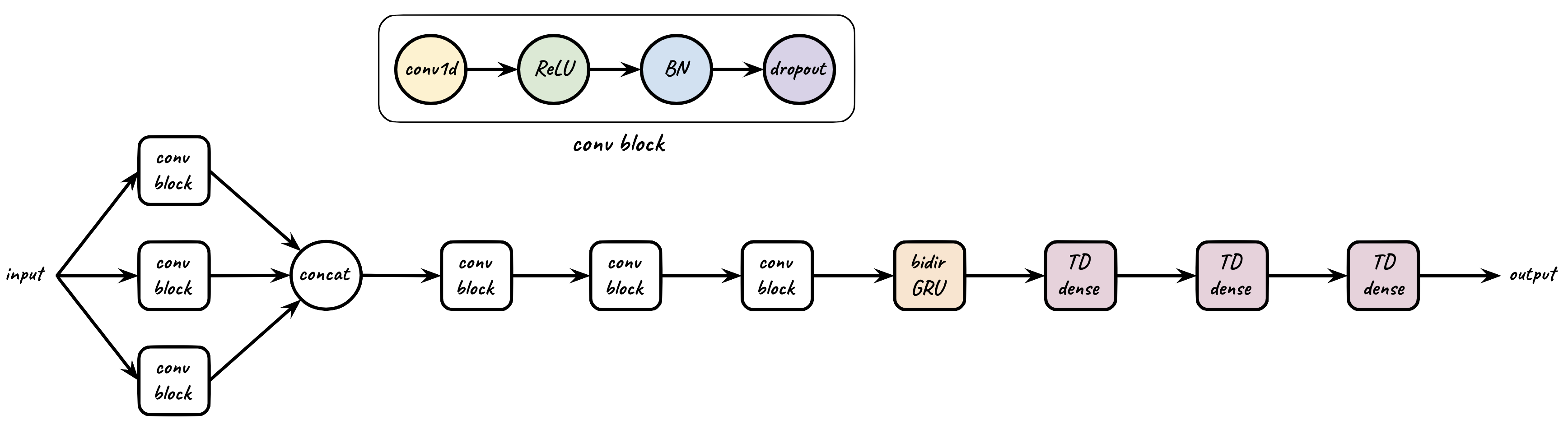}
\caption{Bidirectional GRU with convolutional blocks.}
\label{fig:model1}
\end{figure*}

Figure \ref{fig:model1} shows the architecture for this model. The concatenation of one-hot encoded residue, residue embedding, and residue profiles are passed into multi-scale CNN layers with different kernel sizes (3, 5, 7) to obtain multiple local contextual feature maps \cite{li2016protein}. This is followed by a series of cascading convolutional layers. A series of 3 concatenated 1D convolutions are applied. Each of the convolutions are followed by several layers \cite{chollet2017book}: time distributed ReLU activation, batch normalization and dropout layers (with probability 0.5). This passes through a single (256 unit) bidirectional (CuDNN) GRU \cite{chung2014empirical} with a $l_2$ recurrent regularizer. The output is generated by two fully connected ReLU activated layers (of size 128 and 64) followed by a soft-max output layer.

\vspace{-10pt}

\subsection{Temporal convolutional network}
\begin{figure*}[ht]
\centering
\includegraphics[width=0.8\textwidth]{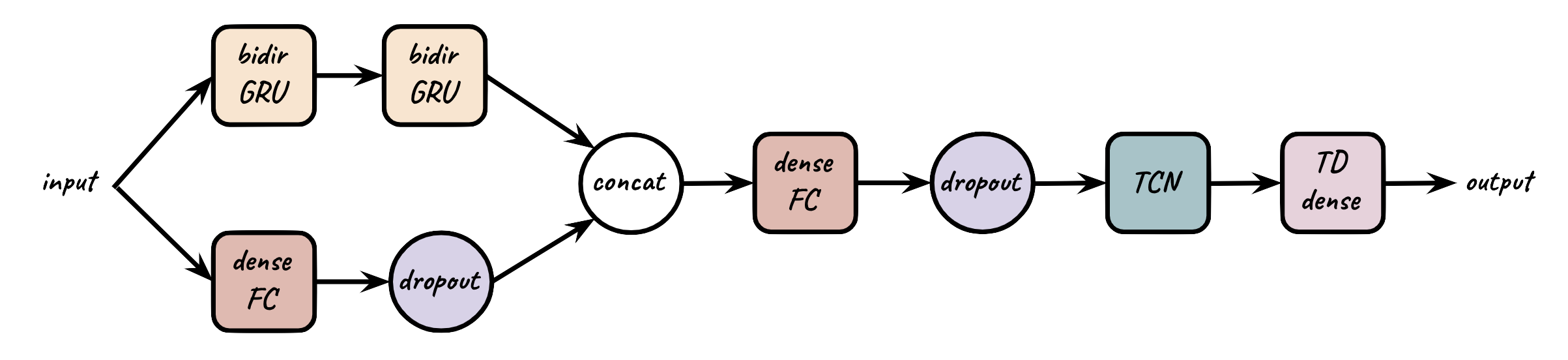}
\caption{Temporal convolutional network (TCN).}
\label{fig:model3}
\end{figure*}

Figure \ref{fig:model3} shows the architecture for this model. Two embedding layers fed with bigrams of the original data are concatenated with profile features. One concatenated output is fed into a dense layer followed by dropout. Another concatenated output is fed into two bidirectional (CuDNN) GRUs. These two, separate layers (the dense and the 2nd bidirectional GRU) are concatenated. The concatenated output is fed into a dense layer, followed by dropout, a temporal convolutional network \cite{van2016wavenet}, and a time-distributed dense layer with softmax activation. 

\vspace{-5pt}

\vspace{-10pt}

\subsection{Bidirectional GRU with 2D convolution}
\begin{figure*}[ht]
\centering
\includegraphics[width=1\textwidth]{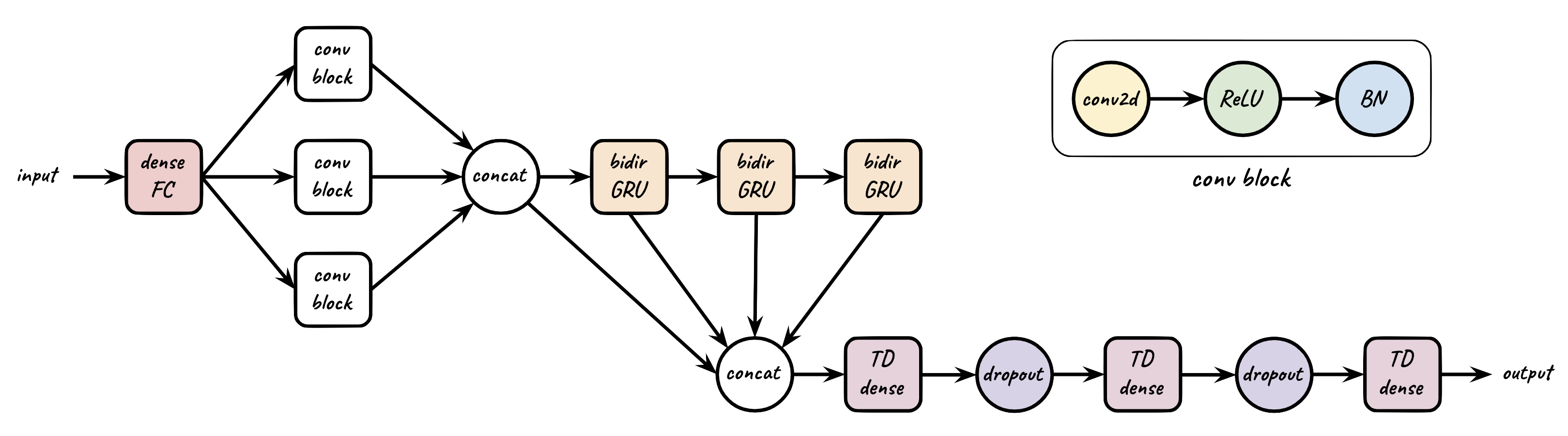}
\caption{Bidirectional GRUs.}
\label{fig:model4}
\end{figure*}

Figure \ref{fig:model4} shows the architecture for this model. The model concatenates the following features as input: a linear combination of the onehot vectors of the preceding amino acids, a linear combination the onehot vectors of the following amino acids, the onehot vector corresponding to the current amino acid and the the profile features for the current amino acid. A fully-connected layer (128 units) removes sparsity from the features, and its outputs are fed into three convolutional layers (3, 7, 11) with 64 filters each. After batch normalization of the outputs, they are concatenated and passed through 3 stacked bidirectional GRUs (with 32 units each). The concatenation of the GRUs' outputs with the convolutional layers' outputs is passed through a two-layer fully connected network. 

\vspace{-5pt}

\vspace{-10pt}

\subsection{Convolutions and bidirectional LSTM}
\begin{figure*}[ht]
\centering
\includegraphics[width=0.9\textwidth]{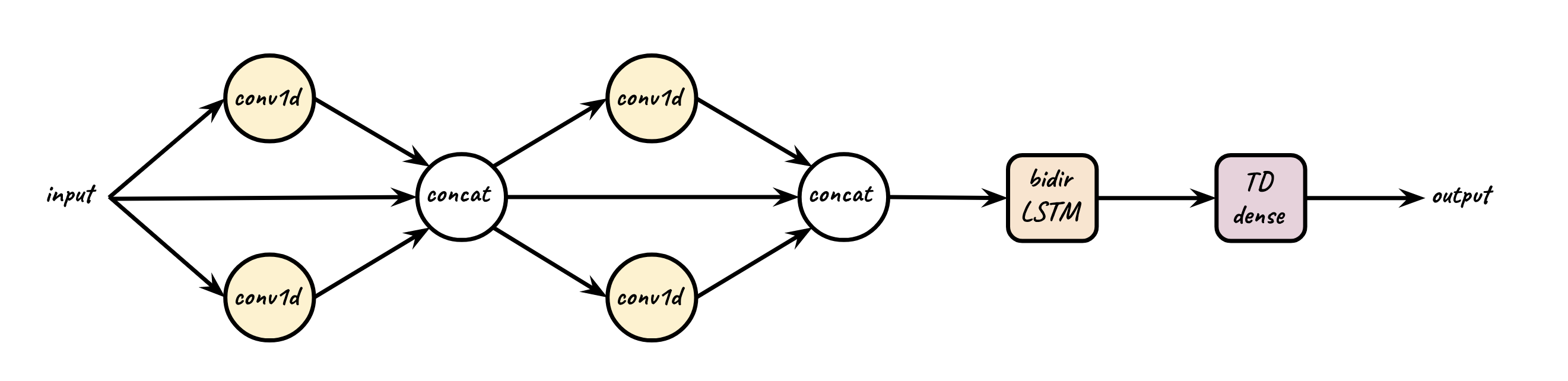}
\caption{Convolutions and bidirectional LSTM.}
\label{fig:model6}
\end{figure*}

Figure \ref{fig:model6} shows the architecture for this model. The model uses skip connections, feeding the encoded input, to two independent convolution layers of 64 channels each (with 11 and 7 kernel sizes respectively). Further, we concatenate both with the input. Now, we again use two independent convolution layers each of 64 channels (with 5 and 3 kernel size respectively). Again, we concatenate the input from the previous concatenation and the output of the two convolution layers. Next, this concatenation is fed to a bidirectional LSTM layer that produces a 128 unit output which is finally used to generate the output using a TD dense layer.

\vspace{-5pt}

\vspace{-10pt}

\subsection{Model hyperparameters}
Table \ref{tab:hyperparams} summarizes the hyperparameters used for each of our models for both of the test sets.

\begin{table}
\small
  \caption{Hyperparameters of each of our models for both of the test sets.}
  \label{tab:hyperparams}
  \centering
  \begin{tabular}{|l|lllll|}
  \hline
    Model & Optimizer & Learning Rate & Decay & Epochs & Batch\\
    \hline
    U-net with convolution blocks & RMSprop & 0.002 & 0.5 & 80 & 128\\
    Bidirectional GRU with conv. blocks & Nadam & 0.002 & 0.004 & 75 & 128\\
    Temporal convolutional network & Adam & 0.001 & 0.0001 & 5 & 16\\
    Bidirectional GRUs & Nadam & 0.002 & 0.004 & 10 & 64\\
    Bidirectional LSTMs with attention & RMSprop & 0.003 & 0.5 & 20 & 64\\
    Convolution and bidirectional LSTM & RMSprop & 0.001 & 0 & 30 & 128\\
    \hline
  \end{tabular}
\end{table}

\section{Results}
\label{results}

\subsection{Unbiased accuracy evaluation using CB513}

Table \ref{tab:cb513comparisonbetweenourmodels} compares mean accuracy between each of our 6 models and their ensemble model. The ensemble is computed by taking the argmax over the average of probabilities over the models for each Q8 structure class, $y = \arg \max_{j} \frac{1}{m}(\sum_{i=1}^{m} p_{i}^{(j)})$, for models $m$ and classes $j$. Table \ref{tab:cb513comparisonwithothers} compares accuracy with other work on the best single model and ensemble for the CB513 dataset. Table \ref{tab:cb513confusionmatrix} shows the confusion matrix for each of the Q8 structures for the CB513 dataset. Table \ref{tab:cb513fscore} shows the precision, recall, and f-score for each of the Q8 structures for the CB513 dataset.

\begin{table}
\small
  \caption{Q8 mean accuracy of our models and their ensemble on the CB513 dataset.}
  \label{tab:cb513comparisonbetweenourmodels}
  \centering
  \begin{tabular}{|l|l|}
    \hline
    Ensemble & 70.7\\
    \hline
    Bidirectional GRU with convolution blocks & 69.8\\
    U-Net with convolution blocks & 69.2\\
    Temporal convolutional network & 68.7\\
    Bidirectional LSTMs with attention & 68.4\\
    Convolutions and bidirectional LSTM & 67.8\\
    Bidirectional GRUs & 67.4\\
    \hline
  \end{tabular}
\end{table}

\begin{table}
\small
  \caption{Q8 mean accuracy using the best single model and ensemble for different methods on CB513 dataset. One apparently relevant study, CRRNN \cite{zhang2018prediction}, which also reports results on CB513, is excluded from the table as its training set is twice as large as the one used by the other methods.}
  \label{tab:cb513comparisonwithothers}
  \centering
  \begin{tabular}{|l|l|l|}
    \hline
    Model & Best Single & Ensemble \\
    \hline
    MUFOLD-SS \cite{fang2018mufold} & 70.5 & 70.6\\
    NCCNN \cite{busia2017next} & 70.3 & 71.4\\
    biRNN-CRF \cite{johansen2017deep} & 69.4 & 70.9\\
    DeepMSCNN \cite{busia2016protein} & 70.0 & 70.6\\
    DCRNN \cite{li2016protein} & 69.4 & 69.7\\
    BLSTM \cite{sonderby2014protein} & 67.4 & N/A\\
    GSN \cite{zhou2014deep} & 66.4 & N/A\\
    DeepCNF \cite{wang2016protein} & N/A & 68.3\\
    Ours & 69.8 & 70.7\\
    \hline
  \end{tabular}
\end{table}

\begin{table}
\small
  \caption{Confusion matrix for each of the Q8 structures for the CB513 dataset. The rows represent the predicted output and the columns represent the ground truth labels.}
  \label{tab:cb513confusionmatrix}
  \centering
\begin{tabular}{|l|llllllll|}
\hline
  & L      & B    & E      & G     & I     & H      & S     & T \\
  \hline
L & 11,828 &  618 & 1,880  &  629  &     4 &    738 & 3,192 & 1,619\\
B &      7 &   31 &    6   &   0   &     0 &      3 & 4     & 0\\
E &  3,167 &  316 & 15,419 &   234 &     2 &    334 & 997   & 565\\
G &    134 &    8 &   24   & 851   &     0 &    233 & 109   & 328\\
I &      0 &    0 &    0   &   0   &     0 &      0 & 0     & 0\\
H &    762 &   77 &  216   & 777   &    22 & 24,126 & 554   & 1,585\\
S &    871 &   49 &  201   &  78   &     0 &     77 & 2,039 & 502\\
T &  1,151 &   82 &  270   & 563   &     2 &    646 & 1,421 & 5,414\\
\hline
\end{tabular}
\end{table}

\begin{table}
\small
  \caption{Precision, recall, and f-scores for each of the Q8 structures for the CB513 dataset.}
  \label{tab:cb513fscore}
  \centering
\begin{tabular}{|l|lll|}
\hline
  & Precision & Recall & F-score\\
\hline
L & 0.58 & 0.66 & 0.62 \\
B & 0.61 & 0.03 & 0.05 \\
E & 0.73 & 0.86 & 0.79 \\
G & 0.50 & 0.27 & 0.35 \\
I & 0.0  & 0.0  & 0.0 \\
H & 0.86 & 0.92 & 0.89 \\
S & 0.53 & 0.25 & 0.34 \\
T & 0.57 & 0.54 & 0.55 \\
\hline
\end{tabular}
\end{table}

\subsection{Accuracy evaluation on CB6133}
For completeness of comparison we provide results on the CB6133 dataset. Table \ref{tab:cb6133comparisonbetweenourmodels} compares mean accuracy between each of our 6 models and their ensemble model. Table \ref{tab:cb6133comparisonwithothers} compares accuracy with other work on the best single model and ensemble. Table \ref{tab:cb6133confusionmatrix} shows the confusion matrix for each of the Q8 structures for the CB6133 dataset. Table \ref{tab:cb6133fscore} shows the precision, recall, and f-score for each of the Q8 structures for the CB6133 dataset.

\begin{table}
\small
  \caption{Q8 mean accuracy of our models and their ensemble on the CB6133 dataset.}
  \label{tab:cb6133comparisonbetweenourmodels}
  \centering
  \begin{tabular}{|l|l|}
    \hline
    Ensemble & 76.3\\
    \hline
    U-Net with convolution blocks & 75.4\\
    Temporal convolutional network & 75.4\\
    Bidirectional GRU with convolution blocks & 74.8\\
    Bidirectional GRUs & 72.9\\
    Convolutions and bidirectional LSTM & 71.6\\
    Bidirectional LSTMs with attention & 68.3\\
    \hline
  \end{tabular}
\end{table}

\begin{table}
\small
  \caption{Q8 mean accuracy using best single model and ensemble for different methods on the CB6133 dataset. Both our best single model 75.4\% and ensemble 76.3\% perform best compared with previously known published methods.}
  \label{tab:cb6133comparisonwithothers}
  \centering
  \begin{tabular}{|l|l|l|}
    \hline
    & Best single  & Ensemble \\
    \hline
    GSN \cite{zhou2014deep} & 72.1 & N/A\\
    DCRNN \cite{li2016protein} & N/A & 73.2\\
    biRNN-CRF \cite{johansen2017deep} & 73.4 & 74.8\\
    CRRNN \cite{zhang2018prediction} & N/A & 74\\
    Ours & 75.4 & 76.3\\
    \hline
  \end{tabular}
\end{table}

\begin{table}[ht]
\small
  \caption{Confusion matrix for each of the Q8 structures for the CB6133 dataset. The rows represent the predicted output and the columns represent the ground truth labels.}
  \label{tab:cb6133confusionmatrix}
  \centering
\begin{tabular}{|l|llllllll|}
\hline
  & L     & B   & E      & G   & I & H      & S     & T \\
\hline
L & 7,218 & 322 & 1,220  & 373 & 0 & 389    & 1,855 & 894 \\
B & 3     & 46  & 17     & 1   & 0 & 1      & 1     & 0 \\
E & 1,445 & 142 & 10,344 & 106 & 0 & 152    & 395   & 233 \\
G & 146   & 5   & 28     & 754 & 0 & 164    & 77    & 209 \\
I & 0     & 0   & 0      & 0   & 0 & 0      & 0     & 0    \\
H & 591   & 34  & 251    & 661 & 0 & 19,085 & 337   & 1,062 \\
S & 406   & 22  & 104    & 37  & 0 & 36     & 1,010 & 165 \\
T & 719   & 55  & 255    & 370 & 0 & 394    & 815   & 3,737\\
\hline
\end{tabular}
\end{table}

\begin{table}
\small
  \caption{Precision, recall and f-scores for each of the Q8 structures for the CB6133 dataset.}
  \label{tab:cb6133fscore}
  \centering
\begin{tabular}{|l|lll|}
\hline
  & Precision & Recall & F-score\\
  \hline
L & 0.58 & 0.68 & 0.63 \\
B & 0.66 & 0.07 & 0.13 \\
E & 0.80 & 0.85 & 0.83 \\
G & 0.54 & 0.33 & 0.41 \\
I & 0.0  & 0.0  & 0.0 \\
H & 0.87 & 0.94 & 0.90 \\
S & 0.59 & 0.23 & 0.32 \\
T & 0.58 & 0.59 & 0.59 \\
\hline
\end{tabular}
\end{table}

\section{Conclusions and future work}
We present new diverse architectures for protein structure prediction, some of which have not been used in the field before, and perform with state of the art accuracy. In future work, these architectures may be used as a starting points for meta learning improved architectures, in a neural architecture search. Finally, in the spirit of reproducible research we make our data, models, and code publicly available \cite{drori18ssp}.

\subsubsection*{Acknowledgments}
We would like to thank the 100 CS/DSI/Stats graduate students at Columbia University of the Fall 2018 Deep Learning course for their participation in an in class protein secondary structure prediction competition. The models which achieved top performance in the competition were invited to participate in this follow-up work, which lead to the discovery of new architectures with state of the art performance. We would like to thank Tomer Sidi of BGU for thorough examination of the correct measures used for performance comparison. We would like to thank Jian Zhou and Olga Troyanskaya of Princeton for making their CB6133 dataset available and for updating their CB6133 dataset splits following our work. Chen Keasar is partially supported by grants 1122/14 from the Israel Science Foundation (ISF).

\clearpage

\small
\bibliography{paper}
\bibliographystyle{plain}

\end{document}